# Improved YOLOv3: Object Classification in Intelligent Transportation System

Yang Zhang [1,3], Changhui Hu [1,2], Xiaobo Lu [1,2,*]

[1] School of Automation, Southeast University, Nanjing 210096, China
[2] Key Laboratory of Measurement and Control of CSE, Ministry of Education, Southeast University
Nanjing 210096, China
[3] Faculty of Engineering and Information Technology, University of Technology Sydney, Ultimo, NSW 2007, Australia
*Corresponding author, E-mail: xblu2013@126.com (X.B.Lu)

**Abstract** The technology of vehicle and driver detection in Intelligent Transportation System(ITS) is a hot topic in recent years. In particular, the driver detection is still a challenging problem which is conductive to supervising traffic order and maintaining public safety. In this paper, an algorithm based on YOLOv3 is proposed to realize the detection and classification of vehicle, driver and people on highway, so as to achieve the purpose of distinguishing driver and passenger and form a one-to-one correspondence between vehicles and drivers. The proposed model and contrast experiment are conducted on our self-build traffic driver's face database. The effectiveness of our proposed algorithm is validated by extensive experiments and verified under various complex highway conditions. Compared with other advanced vehicle and driver detection technologies, the model has a good performance and is robust to road blocking, different attitudes and extreme lighting.

## 1 Introduction

According to the official news of the National Transport Ministry in China, with the rapid rising of vehicles, the number of traffic violation is increasing day by day, which must cause the frequent traffic accident and potential traffic disease. Under this trend, a series of ITS for this tricky issue are put into research. Nowadays, face detection and objective classification technologies are urgently required in this system, such as the application of driver monitoring in Advanced Driver Assistance System(ADAS) [1,2].

Among computer vision area, face detection techniques attract most attention of researchers. With the fast development of digital image processing technology, the frontal faces can be detected and aligned through previous proposed methods [3]. However, when it comes to apply the algorithms into real application, which is affected by various affecting elements such as extreme lighting, varying pose, etc, the robustness of the methods are degraded substantially. As a whole, there exit two main keys affecting face detection in real applications. Firstly, due to the varying visual effect of driver's face among complex backgrounds, an accurate face detector is required to realize the binary classification. Secondly, the face position is difficult to located in the large special scale, meanwhile, the time efficiency is the necessary requirement.

Aiming at adapting to the practical application, in 2004, Viola[4] put forward a well-known architecture combines extracted features and boosted cascade network. The features contain useful information of human face, resulting in fast and accurate detection. At the same time, the cascade network forms an union of the extract features to realize final face vs. non-face classification. The original cascade detection method is based on AdaBoost algorithm with Haar-Like features to perform cascaded classifier. Haar-Like features is conducted to build the binary classifier with accurate performance without redundant computation. At the same time, the cascade network utilizes the extract features to perform classification. Unfortunately, some later researches [5, 6] indicate that, due to the simple innate character of Haar-Like features, it cannot keep continued competitiveness in real applications which affect the visual consistency of faces.

In this condition, more advanced features are suggested to be used in a real-environment detection. Thus, the false positive detection are eliminated efficiently in the early stages. Under this trend, convolutional neural network (CNN) is utilized to extracted advanced features in face detection, which can realizes feature learning among complex environment condition based on extensive training data. Yang et al.[7] proposed the deep neural networks for facial feature recognition, aiming at getting high reply in face regions, resulting in producing candidate windows of human faces. However, they just perform well on some common public dataset constructed under suitable experimental environment. Here, we show some testing example images in FDDB FACE [7] and WIDER FACE[8] in Fig. 1. In addition, these algorithms are time-consuming in real condition.

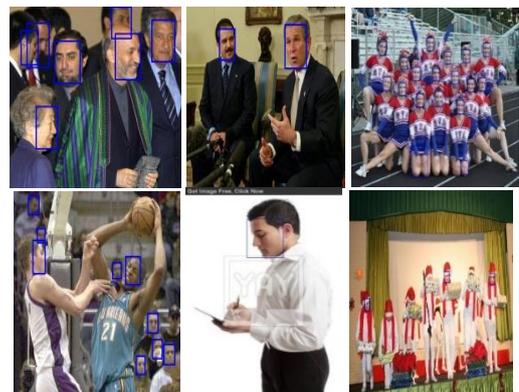

Fig.1. Some testing example images in FDDB and WIDER FACE database.

In this paper, an improved vehicle and driver detection model based on YOLOv3 [11] is proposed, which is called IYOLO to form an adaptive detection model. The model is trained on our own traffic vehicle and driver database, PSD-HIGHROAD database, which is provided by Public Security Department of Jiangsu Province, China, including

---

[a] Corresponding author: xblu2013@126.com

15000 driver's face images captured by the cameras on various highroads in Jiangsu Province. Fig.2 shows some examples of PSD-HIGHROAD database. The experimental results show that compared with other advanced methods, IYOLO model can detect both driver and the vehicle, and classify people in the vehicle, which has a higher detection rate and a lower error detection rate.

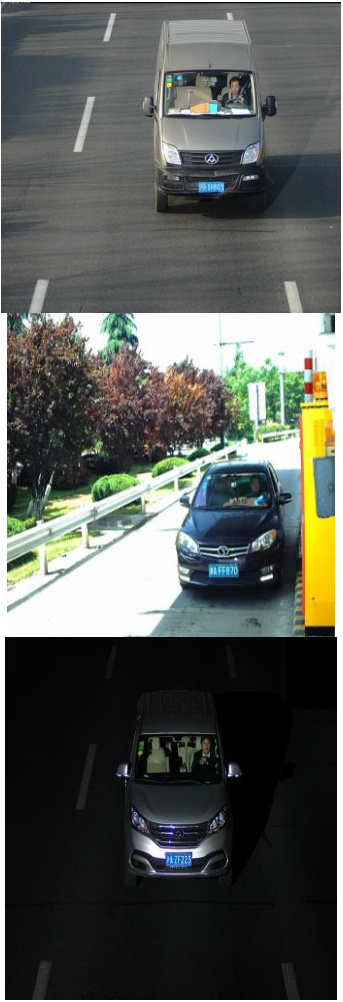

Fig.2. Some example images under varying environment in PSD-HIGHROAD database.

The main contributions of our study can be concluded into four aspects:

a. We provide a particular driver face's database on highroad, PSD-HIGHROAD database, which includes 15000 images under varying environment (morning, afternoon and night) for modelling and detection.

b. We employ the YOLOv3 architecture with receptive fields of different kernel size and propose an advanced regularization method, extracting multi-scale features in image so as to realize adaptive detection and classification in ITS.

c. We utilize the effective technique to realize online hard sample mining based on PSD-HIGHROAD database, improving the model's performance.

d. We conduct extensive experiments on the PSD-HIGHROAD database. Our proposed model shows high detection and classification performance compared to other state-of-the-art method.

## 2 Basic Theory

In this section, we do the research on the basic structure of YOLOv3 and the adaptive regularization method, aiming at realizing adaptive detection and classification in ITS. The improvement on the regularization method and the basic network structure of YOLOv3 used in our model are presented below.

2.1 The egularization Method

Aiming at realizing more effective regularization. This research makes improvement on the original regularization method (cross-entropy loss) used in YOLOv3, adopting label smoothing regularization (LSR) [15].

Recently, Szegedy et al. [15] made discover on the LSR. Briefly, there exits the possibility that the network would be tuned towards the ground truth class, which leads the network to appear over-fitting. Compared to the cross-entropy loss which distributes 0 to the non-ground truth classes directly, LSR uses small values instead.

The cross-entropy loss can be calculated by

$$l = -\sum_{k=1}^{K} \log(p(k)) q(k) \quad (1)$$
$$p(k) \in [0,1] \quad (2)$$

In Eq(1), $k \in \{1,2,...,K\}$ represents the identity of class. p(k) in Eq(2) shows the probability that whether the input sample's predicted class is k, which is outputted by network. It originates from the Softmax function which realizes normalization on the previous fully-connected layer's output. q(k) is the distribution of ground truth, showing below in Eq(3). What should be noticed is that, y is the class label for ground truth.

$$q(k) = \begin{cases} 0 & k \neq y \\ 1 & k = y \end{cases} \quad (3)$$

If we do not take the 0 terms in Eq(1) into consideration, the cross-entropy loss can be calculated only use the ground truth term, showing in Eq(4).

$$l = -\log(p(y)) \quad (4)$$

Thus, aiming at realizing highest predicted probability for the ground-truth class, the effective method is to minimizing the cross-entropy loss. In [15], LSR is proposed to utilize the distribution of the non-ground truth classes, resulting in the less confident for condition that the network points to the ground truth. The $q_{LSR}(k)$, label distribution, is defined below.

$$q_{LSR}(k) = \begin{cases} \frac{\varepsilon}{K} & k \neq y \\ 1 - \varepsilon + \frac{\varepsilon}{K} & k = y \end{cases} \quad (5)$$

In Eq(5), $\varepsilon \in [0,1]$ is a hyperparameter. When $\varepsilon$ is equal to 0, Eq(5) is equal to Eq(3). When the value of $\varepsilon$ is too large, this network will unable to get the ground truth label.

Combining Eq(1) and Eq(5) can derive the modified cross-entropy loss, showing in Eq(6).

$$l_{LSR} = -(1-\varepsilon)\log(p(y)) - \frac{\varepsilon}{K}\sum_{k=1}^{K}\log(p(k)) \quad (6)$$

2.2 YOLOv3

The core idea of YOLOv3 is to use the picture as a network input, directly in the output layer to return to the position of the bounding box and its subordinate categories.

The overall stages of YOLOv3 which is constituted of four parts are illustrated below.

### 2.2.1 Bounding Box Prediction

The anchor boxes of YOLOv3 are made by clustering. The four coordinate values for each of the bounding box prediction ($t_x$, $t_y$, $t_w$, $t_h$) in predicting cell (a picture into S * S grid cells) based on the top left corner of the image offset ($c_x$, $c_y$), as well as the bounding box by the width and height of $p_w$, $p_h$, bounding box can be predicted according to the following way:

$$b_x = \sigma(t_x)+c_x \quad (7)$$

$$b_y = \sigma(t_y)+c_y \quad (8)$$

$$b_w = p_w e^m \quad (9)$$

$$b_h = p_h e^n \quad (10)$$

$$m=t_w , n=t_h \quad (11)$$

Sum of squared error loss is used to predict the coordinate value, so the error can be calculated rapidly.

YOLOv3 predicts the score of an object for each bounding box by logistic regression. If the prediction of the bounding box and the real border overlap better than that of the other all forecasts, then the value is 1. If the overlap does not reach a threshold (setting 0.5), the prediction of bounding box will be ignored, and is displayed as no loss.

### 2.2.2 Class Prediction

To classify different kinds of objections, independent logistic classifiers are used instead of a SoftMax. When training, the regularization method proposed in Section 2.1 is used for the class predictions.

### 2.2.3 Predictions Across Scales

YOLOv3 predicts different boxes in three different scales. YOLOv3 uses FPN (feature pyramid network) to extract feature from scales, and finally predicts a 3-d tensor, containing the bounding box information, object information, and class information.

### 2.2.4 Feature Extractor

YOLOv3 uses a complex network for performing extraction, which has 53 convolutional layers, called Darknert-53. This study simplifies the network structure and makes an effective classification.

## 3 The Proposed Method

### 3.1 The Whole Procedure of IYOLO

Fig.3 performs the procedure chart of the proposed algorithm. Firstly, the IYOLO-Net can get the rough bounding boxes of cars and people in the input image. Secondly, the accurate vehicle and people region are confirmed. Thirdly, driver, passenger and car are given labels. Finally, the picture with bounding boxes and labels is performed.

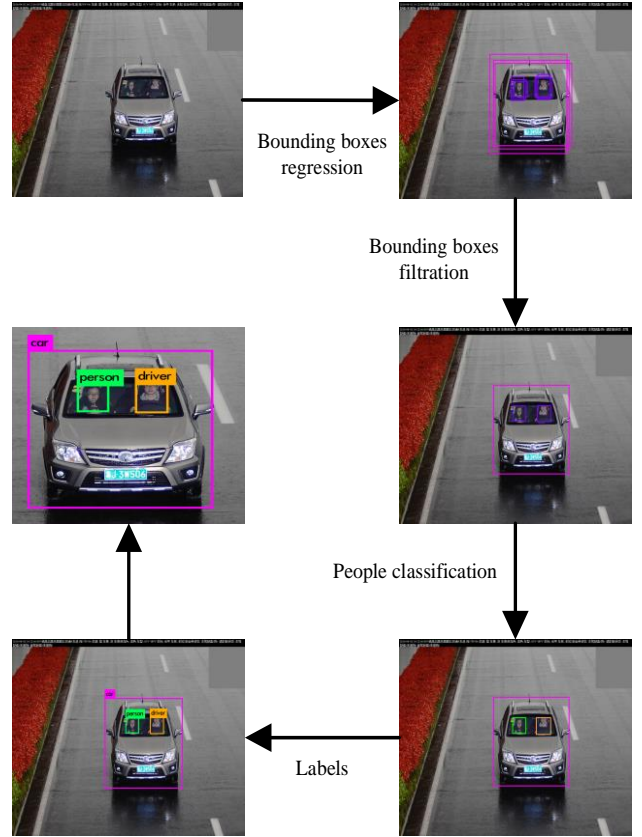

Fig.3. The whole procedure of IYOLO algorithm.

### 3.2 Architecture of Proposed Convolutional Network

In IYOLO model, we make an improvement on the architecture of primary network, which becomes smaller and more efficient (Table 1). The size of input image is 416*416. There are totally four maxpool layers and twenty-two convolutional layers. In the network, the route layer is to bring finer grained features from earlier in the network. The reorg layer is to make these features match the feature map size at the later layer. The end feature map is 13*13, the feature map from earlier is 26*26*512. The reorg layer maps the 26*26*512 feature map onto a 13*13*2048 feature map so that it can be concatenated with the feature maps at 13*13 resolution. By this means, high resolution features and low-resolution features are linked together, which can increase the recognition accuracy of small objects such as our people in the vehicle.

Table 1. The architecture of proposed convolutional network.

| Number | Layer | Filters | Size | Output |
|---|---|---|---|---|
| 0 | convolutional | 32 | 3*3/1 | 416*416 |
| 1 | maxpool | | 2*2/2 | 208*208 |
| 2 | convolutional | 64 | 3*3/1 | 208*208 |
| 3 | maxpool | | 2*2/2 | 104*104 |
| 4 | convolutional | 128 | 3*3/1 | 104*104 |
| 5 | convolutional | 64 | 1*1/1 | 104*104 |
| 6 | convolutional | 128 | 3*3/1 | 104*104 |
| 7 | maxpool | | 2*2/2 | 52*52 |
| 8 | convolutional | 256 | 3*3/1 | 52*52 |
| 9 | convolutional | 128 | 1*1/1 | 52*52 |
| 10 | convolutional | 256 | 3*3/1 | 52*52 |
| 11 | maxpool | | 2*2/2 | 26*26 |
| 12 | convolutional | 512 | 3*3/1 | 26*26 |
| 13 | convolutional | 256 | 1*1/1 | 26*26 |
| 14 | convolutional | 512 | 3*3/1 | 26*26 |
| 15 | convolutional | 256 | 1*1/1 | 26*26 |
| 16 | convolutional | 512 | 3*3/1 | 26*26 |
| 17 | maxpool | | 2*2/2 | 13*13 |
| 18 | convolutional | 1024 | 3*3/1 | 13*13 |
| 19 | convolutional | 512 | 1*1/1 | 13*13 |
| 20 | convolutional | 1024 | 3*3/1 | 13*13 |
| 21 | convolutional | 512 | 1*1/1 | 13*13 |
| 22 | convolutional | 1024 | 3*3/1 | 13*13 |
| 23 | convolutional | 1024 | 3*3/1 | 13*13 |
| 24 | convolutional | 1024 | 3*3/1 | 13*13 |
| 25 | route | | | |
| 26 | reorg | | | |
| 27 | route | | | |
| 28 | convolutional | 1024 | 3*3/1 | 13*13 |
| 29 | convolutional | 40 | 1*1/1 | 13*13 |
| 30 | detection | | | |

### 3.3 Online Hard Sample Mining

We use the training set of PSD-HIGHROAD database to train this model. The testing set is utilized to verify the performance. What should be noticed is that, online hard sample mining is conducted in the training process. In real application, especially in highway condition, as the result of complex environment, it is necessary to generate representative negative sample, strengthening the detector in training procedure. The algorithm computes the loss of all samples among each mini-batch in the forward propagation phase and arrange them from large to small. The top 70% of them are chose as hard samples and the corresponding gradients are calculated to be used in the backward propagation process. Experimental results indicate that this strategy performs better effect, showing in Section 4.

## 4 Experimental Results

In this section, we use our own traffic database, which is constructed by Public Security Department of Jiangsu Province, containing approximately 15000 vehicle images with drivers in different traffic conditions. We randomly choose 12000 images for training IYOLO model, remaining 3000 images for verify the model's performance. We use the open source toolbox Caffe [13] to implement the proposed algorithm based on C++ code, performed on a workstation with Intel Core I7, NVIDIA GTX TITAN X GPU, and the operating system of Ubuntu 16.04. The performances of our method on face detection and classification are showed below.

### 4.1 Calibration Net

According to the requirements of detection and classification, three kinds of training samples are introduced in our training model. IoU (Intersection-over-Union) [18,24] is defined as the coincidence degree between the target window generated by the model and the original labeled face window. If the IoU of region is greater than 0 and less than or equal to 0.3, it is classified as Negative Sample, it is classified as Part Face for the whose IoU is greater than 0.3 and less than or equal to 0.65, and Positive Sample if it is larger than 0.65. The classification model is constructed by negative and positive samples. Then, the bounding box regression is realized by positive samples and part faces. The numbers of various training samples for different objects are shown in Table 2.

Table2. Numbers of various training samples for different objects

| Category | Positive Sample | Negative Sample | Part Face |
|---|---|---|---|
| car | 50,000 | 18,000 | 55,000 |
| person | 45,000 | 20,000 | 55,000 |
| driver | 40,000 | 24,000 | 55,000 |

### 4.2 The Procedure of Training Loss

Fig.4 shows the procedure of training IYOLO model. The picture indicates that the abscissa represents the number of iterations during training and the ordinate represents the loss. When there are objects in the grid, confidence loss of bounding boxes calculates the weight of the contribution to the total loss for five; When there is no object in the grid, confidence loss of bounding boxes calculates the contribution weight to the total loss for one; The weight of the contribution of category loss to the total loss is calculated for one; The weight of the contribution of bounding boxes' coordinates prediction loss to the total loss is calculated for one.

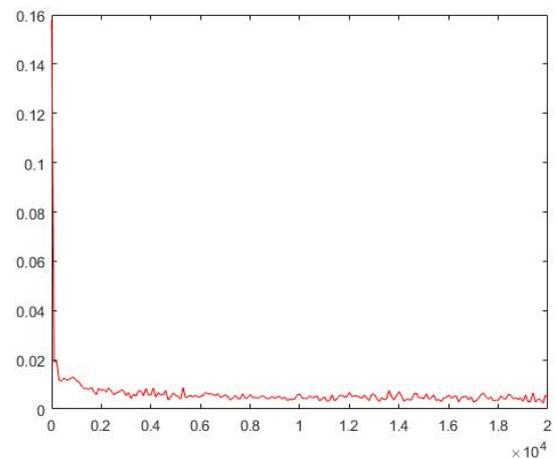

Fig.4. The procedure of training IYOLO model.

### 4.2 The Performances of Detection and Classification

Fig.5 and Fig.6 show several testing samples of IYOLO model under different environment. The detection and classification results of IYOLO model in the dark environment and day time are performed. We can get the conclusion that the proposed model in this paper can realize good performance in complex environment in ITS. Table 3 shows the comparing results, indicating the detecting performances of IYOLO technique with original IYOLO [11] as well as Cascade CNN [19] and Joint Cascade [20]. Fig.7 shows the Precision/Recall curves for our model and other state-of-the-art methods on PSD-HIGHROAD database. In our experiments, the source codes of Cascade CNN as well as Joint Cascade are downloaded in [22] and [23]. The code and processing theory of IoU come from [24]. From the experimental result, the proposed technique in this paper shows high detection rate as well as low error detection rate compared to other state-of-the-art method. This fully proves the IYOLO algorithm owns excellent performance on improving the accuracy on object detection and classification in ITS.

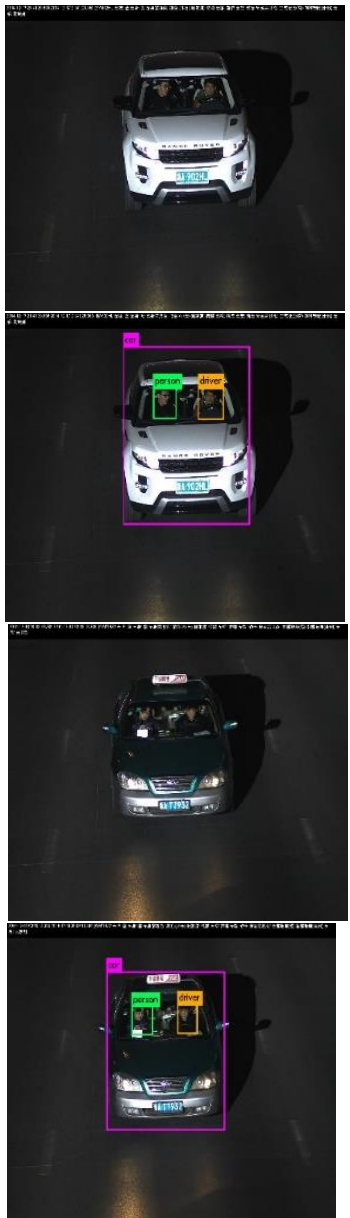

Fig.5. The vehicle and driver detection result of IYOLO model under the dark environment.

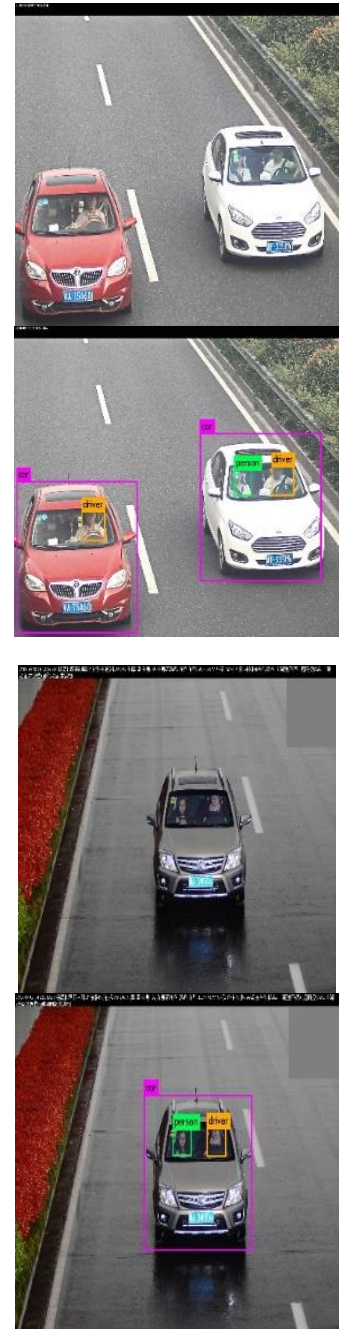

Fig.6. The vehicle and driver detection results of IYOLO model during the day time.

**Table 3.** Detection and classification performances between our method and other comparison techniques.

| Detection Method | IYOLO (OUR) | YOLOv3 [11] | Cascade CNN [19] | Joint Cascade [20] |
|---|---|---|---|---|
| Detection Rate | 92.7% | 85.6% | 78.9% | 71.38% |
| Error Detection Rate | 2.14% | 13.28% | 19.7% | 21.43% |
| Classification Rate | 95% | 81% | NA | NA |
| Error Classification Rate | 0.62% | 6.2% | NA | NA |

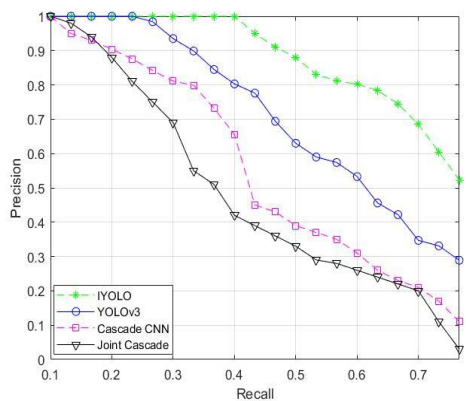

Fig.7. Precision/Recall curves for our model and other state-of-the-art methods on PSD-HIGHROAD database.

### 4.3 The performance of online hard sample mining

Aiming at assessing the effect of online hard sample mining method, we train another model without this strategy in O-net, and compare the corresponding loss curves. The structures and parameters are the same in the two models. Fig.8 shows the results, verifying that the hard sample mining is conductive to improve the method's performance.

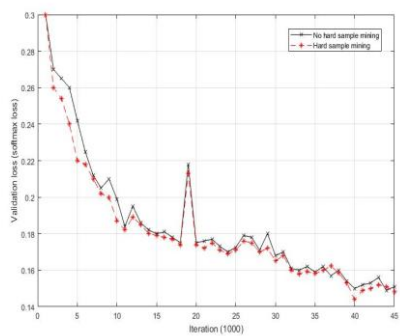

Fig.8. The effect of hard sample mining

### 5 Conclusions

This paper proposes detection and classification method for driver's face on highroad based on Yolov3. This algorithm takes the inherent relationship among face detection and classification into consideration, realizing detection and classification via a coarse-to-fine pattern. What is more, by introducing the PSD-HIGHROAD database into the construction of training model, the robust detection and classification for the face under complex highway environment is realized. Comparative trials on PSD-HIGHROAD database show that IYOLO method owns excellent performance when compared to other similar deep learning methods, such as YOLOv3, Cascade CNN and Joint Cascade. In addition, the proposed method shows fast detecting speed, achieving 14fps on a 8.0GHz CPU. Experimental results indicate that the IYOLO model in this paper is effective, keeping robust to the varying conditions on highroad.

Finally, this research has some limitations under serious illumination environment and shadowed face images. Thus, in future work, denoising method [25-27] is a direction worth studying and making improvements.

### 6 Acknowledgments

This work was supported by the National Natural Science Foundation of China (No.61871123), Key Research and Development Program in Jiangsu Province (No.BE2016739) and a Project Funded by the Priority Academic Program Development of Jiangsu Higher Education Institutions.